\documentclass[letterpaper, 10 pt, conference]{ieeeconf}  

\IEEEoverridecommandlockouts                              

\usepackage[hyphens]{url}
\usepackage{graphicx}
\usepackage{cite}
\usepackage{caption}
\usepackage{xcolor}
\usepackage{algorithm}
\usepackage{algorithmic}
\usepackage{newfloat}
\usepackage{listings}
\usepackage{amsmath,amssymb}
\usepackage{pifont}
\usepackage{adjustbox}
\usepackage{tabularx}
\usepackage{tikz}
\usepackage{booktabs}
\usepackage{multirow}
\usepackage[
    colorlinks=true,    
    urlcolor=blue,      
    linkcolor=blue,     
    citecolor=blue      
]{hyperref}

\DeclareCaptionStyle{ruled}{labelfont=normalfont,labelsep=colon,strut=off}
\floatstyle{ruled}
\newfloat{listing}{tb}{lst}{}
\floatname{listing}{Listing}

\newcommand\lizhou[1]{\textcolor{black}{#1}}

\title{\LARGE \bf
TRIG: Trajectory-Rig Decoupled Metric Geometry Learning
}

\author{Lizhou Liao$^{1,\dagger,*}$, Wentao Xu$^{1,2,\dagger}$, Handong Wang$^{1}$, Lirong Yang$^{1,\ddagger}$, Shuai Yang$^{1}$, Weiwei Liu$^{1}$, Chang Huang$^{1}$%
\thanks{$\dagger$These authors contributed equally to this work.}
\thanks{$^{1}$Carizon. E-mail:
{\tt\small \{lizhou.liao, intern.wentao.xu, handong.wang, lirong.yang, shuai.yang, weiwei.liu, chang.huang\}@carizon.com}.}%
\thanks{$^{2}$ShanghaiTech University. All work was done during the internship.}%
\thanks{$*$Corresponding author.}
\thanks{$\ddagger$Project leader.}
}

\begin{document}

\maketitle
\thispagestyle{empty}
\pagestyle{empty}

\begin{abstract}
\lizhou{Vision-centric autonomous driving requires accurate metric geometry and ego-motion estimation from synchronized multi-camera observations.
Recent visual geometry models show strong performance in pose estimation, depth prediction, and 3D reconstruction, but are not tailored to rigid multi-camera driving systems.
They often encode camera poses as entangled representations, in which time-varying ego-motion and static camera-rig geometry are jointly modeled, limiting the utilization of vehicle-side geometric priors.
We propose \textbf{Trajectory-Rig Decoupled Metric Geometry Learning (TRIG)}, a geometry perception framework for autonomous driving.
TRIG factorizes camera poses into ego-trajectory and camera-rig components, enabling separate modeling of ego-motion and static multi-camera topology.
We introduce decoupled pose encoding and supervision, which separately constrain trajectory evolution and rig geometry for metric-consistent learning.
Moreover, sparse Temporal--Spatial attention separates cross-camera interaction from temporal aggregation, reducing global attention cost while preserving geometric reasoning.
Experiments on five autonomous driving benchmarks show that TRIG achieves state-of-the-art performance in pose estimation, metric depth prediction, and 3D reconstruction.} 
Our project page is at \url{https://slamcabbage.github.io/TRIG/}.
\end{abstract}


\section{Introduction}

\lizhou{Vision-centric autonomous driving has emerged as a scalable and cost-effective paradigm for perception, thereby garnering escalating scholarly and industrial interest~\cite{li2024bevformer,hu2023planning,fu2025orion,gao2024vista,hu2023gaia}. A fundamental challenge in this field is recovering accurate metric scene geometry and ego-motion from synchronized multi-camera observations under rigid camera configurations. Accurate geometric understanding is essential for downstream applications such as planning~\cite{hu2023planning}, simulation~\cite{huang2026s3gaussian}, and online HD map construction~\cite{liao2022maptr}. Recent visual geometry models have demonstrated impressive capabilities in camera pose estimation, depth prediction, and 3D reconstruction from image sequences~\cite{wang2025vggt,yang2025fast3r,wang2025pi}. However, directly applying these models to autonomous driving remains challenging due to rigid multi-camera configurations and the requirement for accurate metric geometry.}

\begin{figure}[!t]
\centering
\includegraphics[width=0.8\columnwidth]{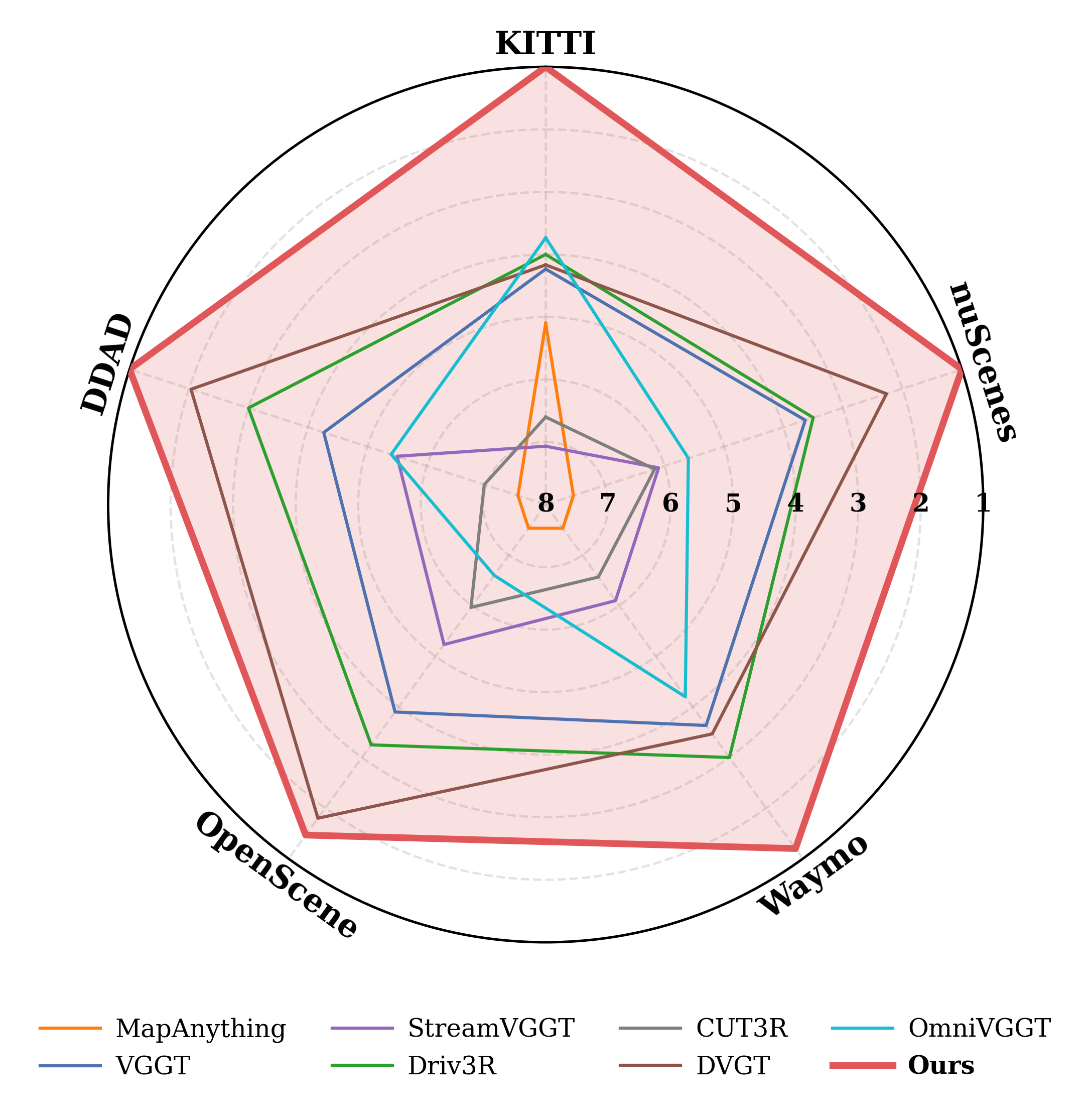}
\caption{\textbf{Radar chart of method rankings across five benchmarks.}
Radial axes represent datasets; rank scores (0--1, higher is better) are averaged over five sub-metrics (Acc, Comp, AbsRel, $\delta_{1.25}$, AUC@$30^\circ$).}
\label{fig:radar}
\end{figure}

Despite substantial progress, reliable metric geometry estimation remains challenging for multi-camera autonomous driving. Existing methods are largely image-driven, using vehicle-side geometry, such as odometry and camera-rig calibration, only as auxiliary inputs or supervision rather than explicit representations for reasoning~\cite{wang2025vggt,jia2025drivevggt,zuo2026dvgt,wang2025pi}. As a result, metric scale is implicitly inferred from visual appearance, causing scale ambiguity and limited geometric accuracy. Recent geometry-prior methods~\cite{peng2026omnivggt,keetha2026mapanything,lin2025depth} are mainly designed for generic visual scenes and treat images as independent observations, thus under-exploiting the structural constraints of synchronized rigid camera rigs.
A camera pose in autonomous driving naturally consists of a time-varying ego trajectory and a static camera-rig configuration. However, existing methods typically entangle them into a unified pose representation, making it difficult to separately exploit motion dynamics and camera topology. Such entanglement further leads to costly global reasoning or post-processing, e.g., $\mathrm{Sim}(3)$ alignment, for consistent metric-scale reconstruction~\cite{zhuo2025streaming,deng2025vggt}.

\begin{figure*}[!t]
    \centering
    \includegraphics[width=0.95\textwidth]{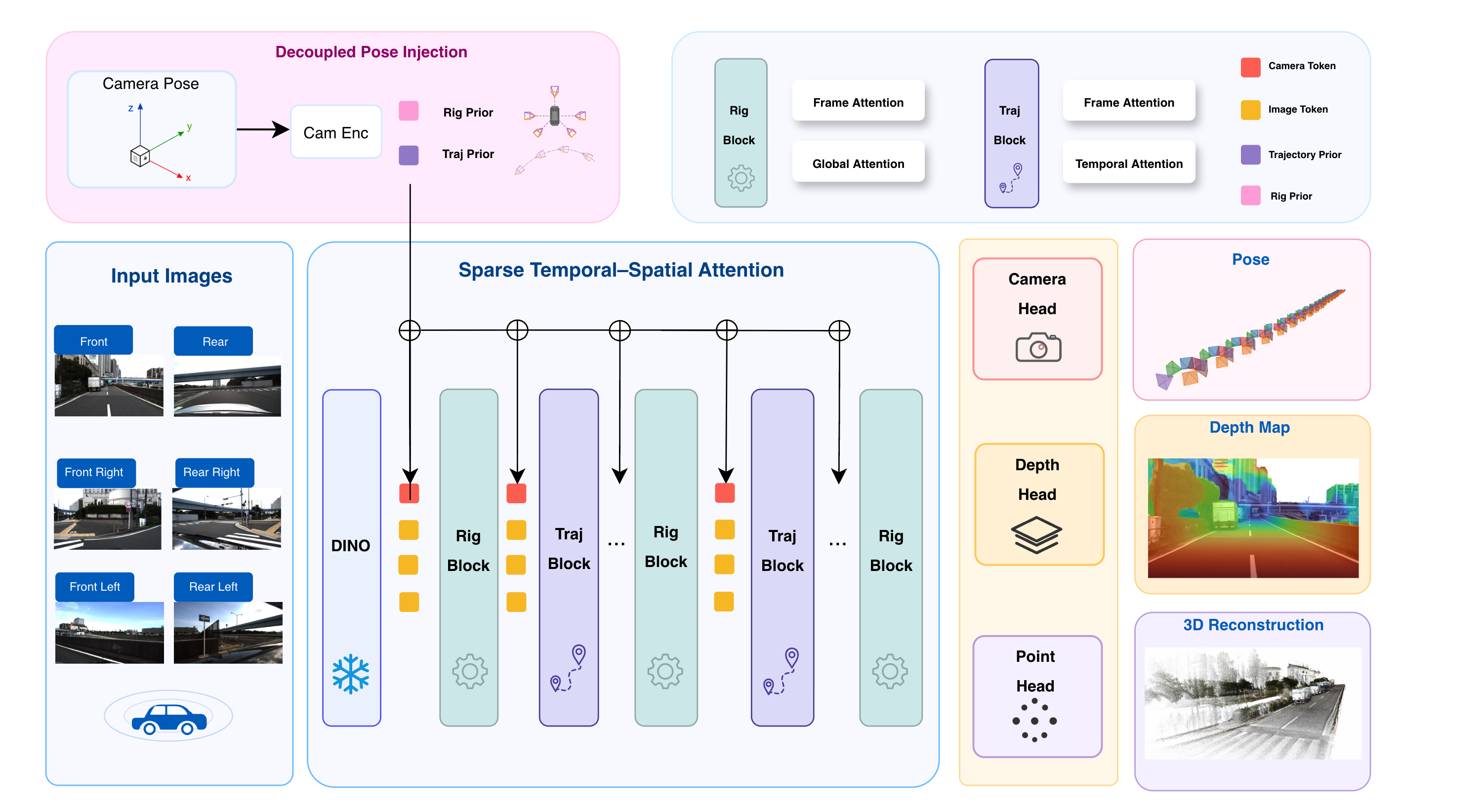}
    \caption{\lizhou{Overview of the proposed TRIG framework. Given synchronized multi-camera images, TRIG decouples pose information into static rig priors and time-varying trajectory priors. Image tokens extracted by a frozen DINO backbone are fused with these priors through sparse Temporal--Spatial Attention. Rig Blocks are sparsely inserted at layers 0, 4, 11, 17, and 23 to inject cross-camera rig constraints, while Traj Blocks capture temporal motion across frames. The original VGGT DPT and camera heads produce the final depth and pose predictions, and are omitted for clarity.}}
    \label{fig:pipeline}
\end{figure*}

Motivated by this observation, we propose \textbf{Trajectory-Rig Decoupled Metric Geometry Learning (TRIG)}, a metric-aware perception framework for multi-camera autonomous driving. TRIG explicitly decouples time-varying ego trajectories from static camera-rig configurations, allowing motion dynamics and camera topology to be modeled independently yet optimized jointly. This decoupling naturally supports an asymmetric temporal--spatial design. The model continuously aggregates trajectory cues but only sparsely injects rig constraints, leading to better scalability and computational efficiency.

Our contributions are three-fold:
\begin{itemize}
    \item We propose \textbf{TRIG}, a trajectory-rig decoupled framework that leverages vehicle-side geometric priors for direct metric-coordinate geometry learning in multi-camera autonomous driving.

    \item We develop a \textbf{decoupled pose modeling strategy}, including pose encoding and pose supervision, to separately handle ego-motion and camera-rig geometry, improving metric scale consistency, pose stability, and global reconstruction quality.

    \item We introduce \textbf{Sparse Temporal--Spatial Attention (STSA)}, which performs continuous trajectory aggregation and sparse rig-constraint injection for scalable cross-camera geometric reasoning. Experiments on five benchmarks show that TRIG achieves state-of-the-art results on metric depth prediction, 3D reconstruction, and ego-pose estimation.
\end{itemize}

\section{Related Work}

\subsection{Visual Geometry Learning for Autonomous Driving}

\lizhou{Vision-based 3D geometry perception has become an essential component of autonomous driving systems, aiming to recover scene structures from multi-camera observations. Early approaches mainly focused on depth estimation from monocular or multi-view images, providing dense depth maps but only partial 3D understanding of the scene geometry \cite{godard2019digging,murez2020atlas,li2023bevdepth}. To achieve a more comprehensive representation, recent methods have explored bird's-eye-view (BEV) perception and 3D occupancy prediction, where the environment is represented in a unified 3D space using voxel-based formulations. These approaches have demonstrated
strong performance
in autonomous driving perception tasks, including semantic occupancy and scene understanding \cite{li2024bevformer, huang2021bevdet, tian2023occ3d}.}

\lizhou{However, voxel-based representations suffer from discretization errors and limited spatial resolution, which makes it challenging to accurately model fine-grained geometric structures. Recently, general visual geometry models have advanced 3D reconstruction by predicting dense point maps from image sequences~\cite{wang2024dust3r,wang2025vggt,yang2025fast3r,zuo2026dvgt,jia2025drivevggt}. Despite their impressive reconstruction capability, most existing approaches estimate geometry in relative coordinate systems and require additional scale recovery~\cite{wang2024dust3r,wang2025vggt,wang2025pi,zuo2026dvgt,jia2025drivevggt,peng2026omnivggt}. }
\lizhou{In contrast, TRIG focuses on directly learning metric geometry for rigid multi-camera driving systems by explicitly leveraging vehicle-side geometric priors.}

\subsection{Geometry-aware Visual Models with Camera Priors}

\lizhou{Existing autonomous driving perception methods commonly exploit calibrated camera geometry through explicit projection using camera intrinsics and extrinsics.}
\lizhou{Conventional autonomous driving perception methods typically rely on explicit camera calibration and geometric projection, such as transforming image features into BEV space using intrinsics and extrinsics parameters~\cite{li2024bevformer,li2023bevdepth}. While effective under fixed camera configurations, these approaches often depend heavily on camera-specific assumptions, limiting their generalization to different sensor setups~\cite{peng2023bevsegformer}.}

\lizhou{Recent visual geometry models have explored learning camera poses, depth, and 3D structures jointly from image sequences~\cite{wang2024dust3r,wang2025vggt,yang2025fast3r,wang2025pi}. Some approaches further incorporate geometric priors, such as camera poses, depth constraints, or external measurements, to improve reconstruction quality~\cite{peng2026omnivggt,keetha2026mapanything,li2025rig3r}. However, most existing methods treat camera poses as a unified geometric condition, without explicitly separating different sources of geometric variation. In multi-camera autonomous driving systems, camera poses naturally contain both time-varying ego-motion trajectories and camera-rig configurations, which provide complementary geometric information. TRIG explicitly decouples these two factors to better exploit vehicle-side priors for metric geometry learning.}

\lizhou{Moreover, existing visual geometry models commonly rely on global self-attention to capture interactions among multi-view and multi-frame inputs, resulting in quadratic computational complexity with increasing input size. Recent works explore efficient attention or streaming strategies to improve scalability~\cite{zhuo2025streaming,deng2025vggt,jia2025drivevggt,chen2026geometric}. Different from these approaches, TRIG introduces a sparse temporal--spatial attention mechanism that separates cross-camera spatial interaction from temporal feature aggregation, reducing computational overhead while maintaining effective geometric reasoning.}

\section{Preliminary}

\lizhou{We adopt VGGT~\cite{wang2025vggt} as the basic visual geometry foundation model, which predicts camera parameters and dense scene geometry from multi-view images in a feed-forward manner. Given an image set $\mathcal{I}$, a visual encoder first extracts image features and converts them into spatial tokens $\mathbf{f}$. These tokens are combined with learnable camera tokens $\mathbf{c}$ and register tokens $\mathbf{r}$ to form the input representation of the transformer encoder:}

\begin{equation}
(\hat{\mathbf{c}}, \hat{\mathbf{r}}, \hat{\mathbf{f}})=E(\mathbf{c},\mathbf{r},\mathbf{f}).
\end{equation}

\lizhou{The encoder is constructed upon the Alternating Attention (AA) mechanism, which performs local and global feature interactions to capture both intra-view structures and cross-view geometric relationships. After several AA blocks, the refined spatial tokens are decoded by dense prediction heads to estimate depth maps and 3D point maps, while the camera tokens are used to regress camera intrinsics and extrinsics. Through this unified architecture, VGGT achieves joint estimation of scene geometry and camera motion from image sequences.}

\begin{figure*}[t]
\centering
\begin{tikzpicture}
    \node[anchor=south west, inner sep=0] (image) at (0,0) {\includegraphics[width=0.94\textwidth]{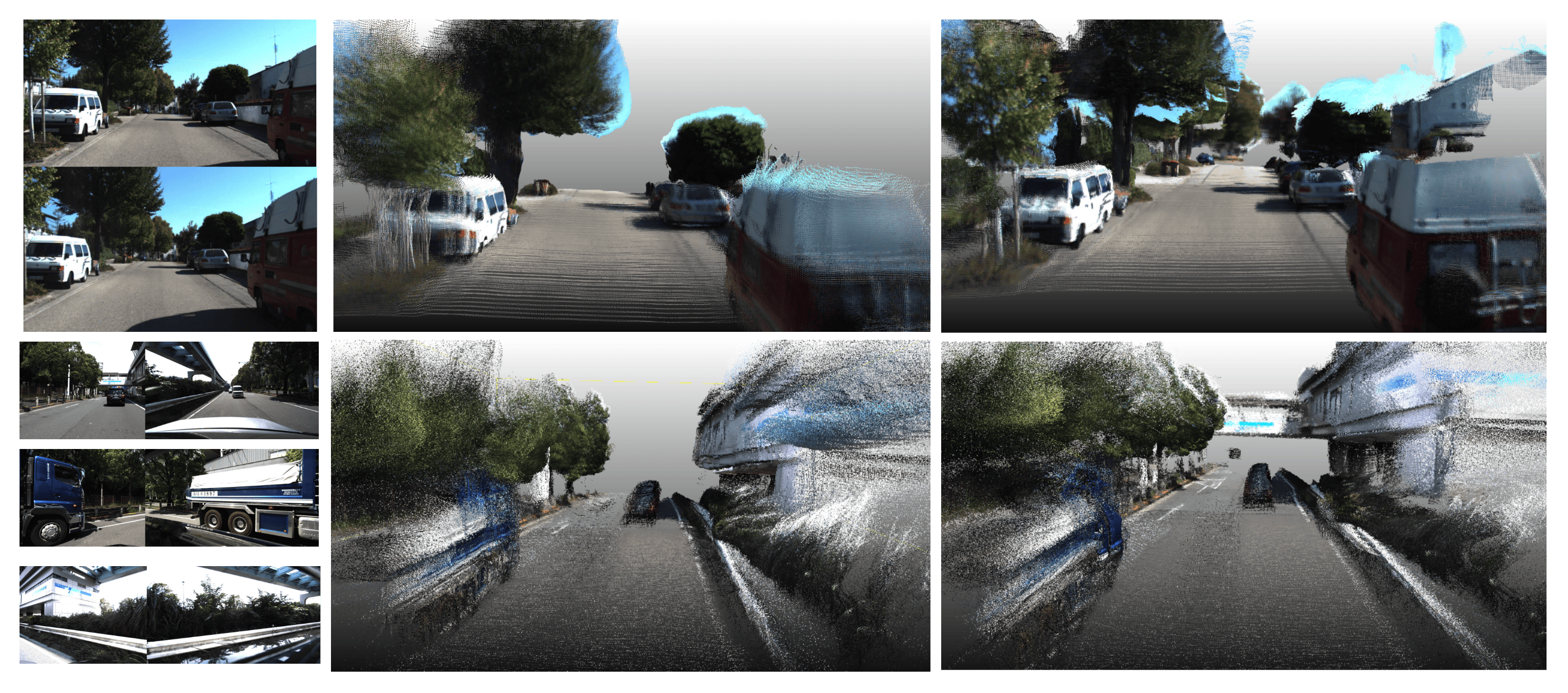}};

    \begin{scope}[x={(image.south east)},y={(image.north west)}]
        \node[anchor=south, font=\bfseries\small, yshift=5pt] at (0.10, 1.0) {Inputs};
        \node[anchor=south, font=\bfseries\small, yshift=5pt] at (0.40, 1.0) {DVGT};
        \node[anchor=south, font=\bfseries\small, yshift=5pt] at (0.80, 1.0) {TRIG (Ours)};

        \node[anchor=east, font=\bfseries\small, xshift=-5pt] at (0.0, 0.75) {KITTI};
        \node[anchor=east, font=\bfseries\small, xshift=-5pt] at (0.0, 0.25) {DDAD};
    \end{scope}
\end{tikzpicture}
\caption{\textbf{Qualitative comparison of 3D reconstruction.}
TRIG produces sharper geometry, cleaner object boundaries, and more consistent large-scale scene structure than DVGT across representative driving scenes.}
\label{fig:qualitative_results}
\end{figure*}
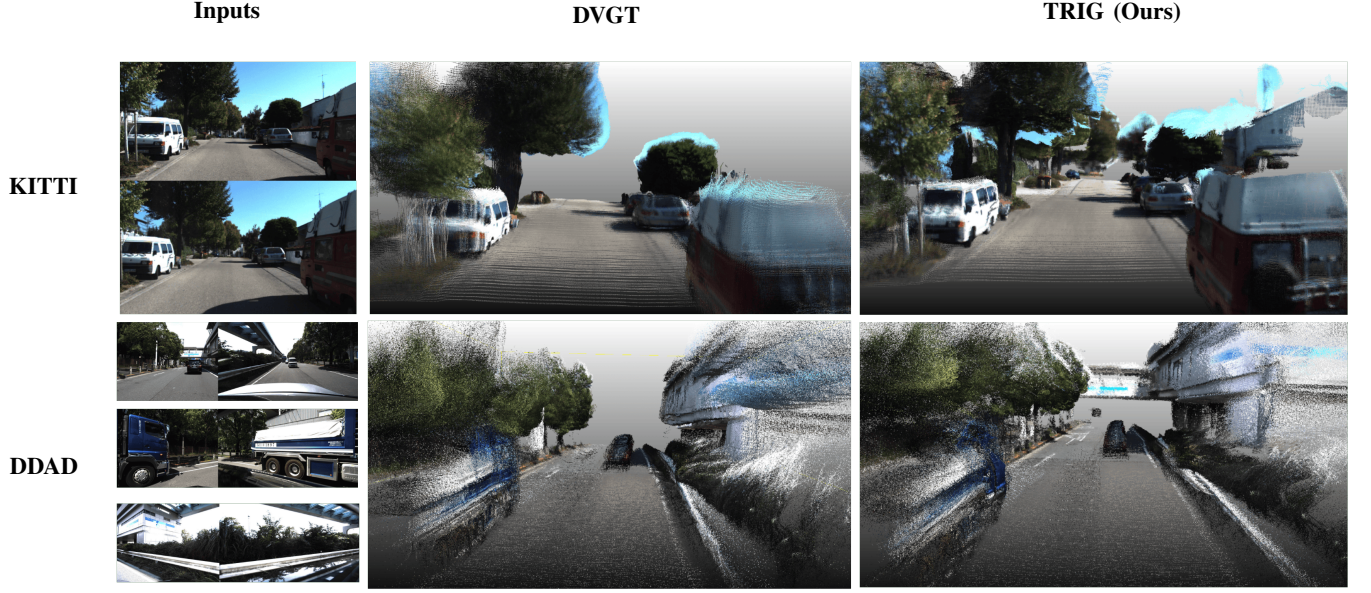

\section{Methodology}

\lizhou{This section details \textbf{TRIG}, a trajectory-rig decoupled framework for multi-camera geometry learning: 1) decoupled pose encoding for metric-aware geometry learning, 2) sparse temporal--spatial attention for efficient long-horizon reconstruction, and 3) decoupled pose supervision for decoupling motion and rig constraints during training.}

\subsection{Overview of TRIG}

\lizhou{TRIG targets multi-camera geometry learning from a temporal sequence of synchronized surround-view images. Given an input clip $\mathcal{I}=\{I_t^c \mid t=1,\dots,T,\ c=1,\dots,C\}$, where $T$ and $C$ denote the number of frames and cameras, respectively, each image $I_t^c\in\mathbb{R}^{3\times H\times W}$ is associated with a camera pose $P_{w \leftarrow c}^{t}\in SE(3)$. Here, $P_{w \leftarrow c}^{t}$ denotes the pose of camera $c$ in the world coordinate system at time $t$. Instead of using $P_{w \leftarrow c}^{t}$ as a single entangled condition, TRIG factorizes it into an ego-trajectory component and a camera-rig component:}

\begin{equation}
P_{w \leftarrow c}^{t} = P_{w \leftarrow \mathrm{ego}}^{t} P_{\mathrm{ego} \leftarrow c}^{c} ,
\label{eq:pose_transform}
\end{equation}

\lizhou{where $P_{w \leftarrow ego}^{t}\in SE(3)$ is the ego-vehicle pose in the world coordinate system at time $t$, and $P_{ego \leftarrow c}^{c}\in SE(3)$ is the fixed extrinsic transform from camera $c$ to the ego-vehicle coordinate system.}

\lizhou{As shown in Fig.~\ref{fig:pipeline}, the pose factors are encoded into trajectory and rig priors,}
\begin{equation}
z_{\mathrm{traj}}^t = \phi_{\mathrm{traj}}(P_{w \leftarrow \mathrm{ego}}^{t}),
\qquad
z_{\mathrm{rig}}^c = \phi_{\mathrm{rig}}(P_{\mathrm{ego} \leftarrow c}^{c} )
\end{equation}

\lizhou{and injected into the image tokens extracted by a frozen visual backbone. TRIG alternates Rig Blocks and Traj Blocks to separately model cross-camera spatial constraints and cross-frame temporal motion cues. This decoupled design enables sparse temporal--spatial interaction over the token set}
\begin{equation}
\mathcal{X} = \{x_t^c\} \cup \{z_{\mathrm{rig}}^c\} \cup \{z_{\mathrm{traj}}^t\}.
\end{equation}
\lizhou{where $x_t^c$ denotes the image tokens of view $c$ at time $t$. The resulting features $ \mathcal{X} $ are fed into the original VGGT DPT head and camera head for depth and camera prediction, respectively. These heads are kept unchanged and thus omitted in Fig.~\ref{fig:pipeline} for clarity.}

\subsection{Decoupled Pose Encoding}

\lizhou{TRIG factorizes multi-camera poses into two complementary components: a temporal ego-motion trajectory and a static camera-rig configuration.
Given the input pose sequence $\mathbf{P}\in\mathbb{R}^{T\times N_C\times 9}$, we use the reference camera to represent the shared trajectory, i.e., $\mathbf{P}_{\mathrm{traj}}=\{P^{t,0}\}_{t=1}^{T}$, and encode it as
$z_{\mathrm{traj}}^{t}=\mathrm{MLP}_{\mathrm{traj}}(P^{t,0})$.
Similarly, the camera rig is extracted from the reference timestamp as
$\mathbf{P}_{\mathrm{rig}}=\{P^{0,c}\}_{c=1}^{N_C}$, and each camera-specific configuration is encoded by
$z_{\mathrm{rig}}^{c}=\mathrm{MLP}_{\mathrm{rig}}(P^{0,c})$.
The final geometry-aware pose embedding is obtained by additive composition:
\begin{equation}
z_{\mathrm{geo}}^{t,c}=z_{\mathrm{traj}}^{t}+z_{\mathrm{rig}}^{c}.
\end{equation}
This decomposition separates dynamic ego-motion from static multi-camera geometry, yielding structured geometric priors for surround-view reasoning.}

\lizhou{Instead of replacing the original VGGT camera tokens, we inject the decoupled pose representation through residual adaptation.
Specifically, the geometry feature is projected by a lightweight adapter and added to the original camera token:
\begin{equation}
\tilde{z}_{\mathrm{cam}}^{t,c}
=
z_{\mathrm{cam}}^{t,c}
+
\mathrm{MLP}_{\mathrm{adapter}}(z_{\mathrm{geo}}^{t,c}),
\end{equation}
where $z_{\mathrm{cam}}^{t,c}$ denotes the camera token produced by VGGT.
Following OmniVGGT~\cite{peng2026omnivggt}, the adapted camera tokens are inserted into each frame attention block during transformer refinement, enabling progressive geometric conditioning while preserving the original visual representation.}

\begin{table*}[t]
\centering
\caption{\textbf{Quantitative 3D reconstruction results across diverse datasets.}
The ``Metric'' column indicates whether a method predicts metric scale directly (\ding{51}) or requires post-hoc Sim(3) alignment with sparse LiDAR depth via the Umeyama~\cite{umeyama1991least} algorithm to recover the metric scale (\ding{55}).
Unlike prior methods, our TRIG predicts metric scale directly and requires no post-hoc alignment.
The top-2 results are highlighted as \textbf{first} and \underline{second}.}
\label{tab:recon}
\setlength{\tabcolsep}{3pt}
\renewcommand{\arraystretch}{0.9}
\footnotesize
\begin{tabular*}{\textwidth}{@{\extracolsep{\fill}}l c cccccccccc@{}}
\toprule
\multirow{2}{*}{Method} & \multirow{2}{*}{Metric}
& \multicolumn{2}{c}{KITTI}
& \multicolumn{2}{c}{NuScenes}
& \multicolumn{2}{c}{Waymo}
& \multicolumn{2}{c}{OpenScene}
& \multicolumn{2}{c}{DDAD} \\
\cmidrule(lr){3-4}\cmidrule(lr){5-6}\cmidrule(lr){7-8}\cmidrule(lr){9-10}\cmidrule(lr){11-12}
& & Acc\,$\downarrow$ & Comp\,$\downarrow$
& Acc\,$\downarrow$ & Comp\,$\downarrow$
& Acc\,$\downarrow$ & Comp\,$\downarrow$
& Acc\,$\downarrow$ & Comp\,$\downarrow$
& Acc\,$\downarrow$ & Comp\,$\downarrow$ \\
\midrule
OmniVGGT~\cite{peng2026omnivggt}        & \ding{55} & 1.111 & 1.116 & 1.310 & 1.703 & 1.443 & 1.423 & 2.704 & 3.598 & 2.535 & 4.317 \\
CUT3R~\cite{wang2025continuous}         & \ding{55} & 0.965 & 2.050 & 2.054 & 2.603 & 3.391 & 4.216 & 1.864 & 2.258 & 2.774 & 4.677 \\
VGGT~\cite{wang2025vggt}                & \ding{55} & 1.154 & 1.294 & 1.300 & 1.498 & 1.641 & 2.053 & 1.422 & 1.496 & 1.741 & 2.473 \\
MapAnything~\cite{keetha2026mapanything} & \ding{51} & 1.880 & \underline{1.014} & 4.499 & 4.886 & 10.205 & 8.494 & 3.353 & 4.303 & 8.015 & 8.493 \\
StreamVGGT~\cite{zhuo2025streaming}     & \ding{55} & 3.421 & 2.196 & 2.588 & 2.414 & 3.630 & 3.275 & 2.304 & 2.098 & 2.717 & 2.788 \\
Driv3R~\cite{fei2024driv3r}             & \ding{55} & 0.864 & 1.083 & 0.742 & 1.345 & \underline{0.800} & \underline{1.311} & 0.884 & 1.693 & 0.950 & 1.259 \\
DVGT~\cite{zuo2026dvgt}                 & \ding{51} & \underline{0.846} & 1.468 & \underline{0.457} & \underline{0.494} & 1.714 & 2.216 & \underline{0.402} & \underline{0.481} & \underline{0.751} & \underline{1.009} \\
\midrule
\textbf{TRIG (Ours)} & \ding{51} & \textbf{0.333} & \textbf{0.351} & \textbf{0.314} & \textbf{0.361} & \textbf{0.509} & \textbf{1.192} & \textbf{0.380} & \textbf{0.434} & \textbf{0.587} & \textbf{0.842} \\
\bottomrule
\end{tabular*}
\end{table*}

\begin{table*}[t]
\centering
\caption{\textbf{Quantitative metric depth results across diverse datasets.}
We report the Absolute Relative Error (Abs Rel $\downarrow$) and the inlier ratio $\delta_{1.25}$ ($\uparrow$), where $\delta_{1.25}$ denotes the percentage of pixels with $\delta\!<\!1.25$.
The top-2 results are highlighted as \textbf{first} and \underline{second}.}
\label{tab:depth}
\setlength{\tabcolsep}{3pt}
\renewcommand{\arraystretch}{0.9}
\footnotesize
\begin{tabular*}{\textwidth}{@{\extracolsep{\fill}}l cccccccccc@{}}
\toprule
\multirow{2}{*}{Method}
& \multicolumn{2}{c}{KITTI}
& \multicolumn{2}{c}{NuScenes}
& \multicolumn{2}{c}{Waymo}
& \multicolumn{2}{c}{OpenScene}
& \multicolumn{2}{c}{DDAD} \\
\cmidrule(lr){2-3}\cmidrule(lr){4-5}\cmidrule(lr){6-7}\cmidrule(lr){8-9}\cmidrule(lr){10-11}
& AbsRel\,$\downarrow$ & $\delta_{1.25}\,\uparrow$
& AbsRel\,$\downarrow$ & $\delta_{1.25}\,\uparrow$
& AbsRel\,$\downarrow$ & $\delta_{1.25}\,\uparrow$
& AbsRel\,$\downarrow$ & $\delta_{1.25}\,\uparrow$
& AbsRel\,$\downarrow$ & $\delta_{1.25}\,\uparrow$ \\
\midrule
OmniVGGT~\cite{peng2026omnivggt}        & \underline{0.067} & \underline{0.949} & 0.175 & 0.440 & \underline{0.096} & 0.737 & 0.304 & 0.655 & 0.225 & 0.591 \\
CUT3R~\cite{wang2025continuous}         & 0.217 & 0.659 & 0.332 & 0.547 & 0.291 & 0.562 & 0.278 & 0.593 & 0.870 & 0.315 \\
VGGT~\cite{wang2025vggt}                & 0.158 & 0.801 & 0.243 & 0.729 & 0.176 & 0.811 & 0.241 & 0.719 & 0.613 & 0.476 \\
MapAnything~\cite{keetha2026mapanything} & 0.188 & 0.725 & 0.568 & 0.269 & 0.507 & 0.211 & 0.486 & 0.240 & 1.971 & 0.195 \\
StreamVGGT~\cite{zhuo2025streaming}     & 0.362 & 0.469 & 0.412 & 0.540 & 0.339 & 0.584 & 0.319 & 0.607 & 0.838 & 0.415 \\
Driv3R~\cite{fei2024driv3r}             & 0.164 & 0.784 & 0.189 & 0.721 & 0.168 & 0.770 & 0.188 & 0.740 & 0.185 & 0.740 \\
DVGT~\cite{zuo2026dvgt}                 & 0.136 & 0.849 & \underline{0.069} & \underline{0.953} & 0.106 & \textbf{0.921} & \underline{0.049} & \underline{0.971} & \underline{0.152} & \underline{0.837} \\
\midrule
\textbf{TRIG (Ours)} & \textbf{0.046} & \textbf{0.967} & \textbf{0.051} & \textbf{0.974} & \textbf{0.094} & \underline{0.889} & \textbf{0.041} & \textbf{0.977} & \textbf{0.109} & \textbf{0.903} \\
\bottomrule
\end{tabular*}
\end{table*}

\subsection{Sparse Temporal--Spatial Attention}

\lizhou{Given the decoupled pose tokens, TRIG introduces a sparse temporal--spatial attention mechanism to aggregate geometric cues from long-horizon synchronized multi-camera sequences.
Let $\mathcal{X}\in\mathbb{R}^{B\times S\times P\times D}$ denote the input token tensor, where $S=T\times N_C$, $P$ is the number of tokens per observation, and $D$ is the token dimension.
Each observation contains image tokens and the injected decoupled pose tokens.
In contrast to applying dense self-attention over all tokens at every layer, TRIG factorizes token interactions into three structured patterns: frame attention, global attention, and temporal attention, which respectively model local visual refinement, cross-camera communication, and ego-motion consistency.}

\paragraph{Structured attention.}
\lizhou{Frame attention performs self-attention independently within each camera observation by reshaping $\mathcal{X}$ into $(BS)\times P\times D$.
This operation refines local image and pose tokens without mixing information across different cameras or timestamps.
Global attention flattens all observations and tokens into $B\times (SP)\times D$, enabling communication across all frames and cameras.
It provides the interaction path required for cross-camera geometric consistency.
Temporal attention first reshapes $\mathcal{X}$ into $B\times T\times N_C\times P\times D$, and then applies self-attention along each camera stream with shape $(BN_C)\times (TP)\times D$. This design aggregates long-range temporal cues while preserving camera-specific viewing directions.}

\paragraph{Rig and trajectory blocks.}
\lizhou{Based on these attention patterns, we define two structured blocks.
The Rig Block applies frame attention followed by global attention, i.e.,
$\mathrm{RigBlock}(\mathcal{X})=\mathrm{GlobalAttn}(\mathrm{FrameAttn}(\mathcal{X}))$.
Since the token set contains rig-aware pose tokens, the global interaction is conditioned on the fixed camera layout, encouraging cross-view geometric consistency.
The Traj Block applies frame attention followed by temporal attention, i.e.,
$\mathrm{TrajBlock}(\mathcal{X})=\mathrm{TempAttn}(\mathrm{FrameAttn}(\mathcal{X}))$.
Unlike the Rig Block, the Traj Block constrains long-range aggregation to intra-camera observations, enabling temporal reasoning under ego-motion while preserving view-specific geometry.}

\paragraph{Sparse temporal--spatial composition.}
\lizhou{The transformer employs Traj Blocks across all layers to facilitate continuous temporal modeling, while sparsely interleaving Rig Blocks to enable global multi-camera interaction.
\lizhou{
Following the design of VGGT\cite{deng2025vggt}, we inject Rig Blocks at indices $\mathcal{L}_{\mathrm{rig}} = \{0, 4, 11, 17, 23\}$ within our 24-layer transformer. The layer update is formulated as:
\begin{equation}
\mathcal{X}^{l+1}
=
\begin{cases}
\mathrm{RigBlock}^{l}\bigl(\mathrm{TrajBlock}^{l}(\mathcal{X}^{l})\bigr),
& l\in\mathcal{L}_{\mathrm{rig}}, \\
\mathrm{TrajBlock}^{l}(\mathcal{X}^{l}),
& l\notin\mathcal{L}_{\mathrm{rig}},
\end{cases}
\end{equation}
This sparse composition effectively decouples cross-camera spatial interaction from temporal feature aggregation, preserving essential geometric information flow while avoiding the prohibitive computational cost of global attention at every layer.
}
Overall, this sparse composition separates cross-camera spatial interaction from temporal feature aggregation, thereby preserving essential geometric interaction paths while reducing unnecessary dense token interactions.}

\subsection{Decoupled Pose Supervision}

\lizhou{To align the training objective with the trajectory-rig factorization of TRIG, we supervise relative poses instead of absolute camera poses.
Relative transformations are invariant to the global coordinate frame and directly capture geometric consistency among observations.
We divide relative pose pairs into two groups: cross-camera pairs at the same timestamp for rig supervision, and cross-time pairs within the same camera stream for trajectory supervision.}

\begin{table}[t]
\centering
\caption{\textbf{Quantitative camera pose results across diverse datasets.} The top-2 results are highlighted as \textbf{first} and \underline{second}.}
\label{tab:pose}
\setlength{\abovecaptionskip}{3pt}
\setlength{\belowcaptionskip}{3pt}
\setlength{\tabcolsep}{3pt}\renewcommand{\arraystretch}{0.85}\footnotesize
\begin{tabular}{@{}l ccccc@{}}
\toprule
Method & \multicolumn{5}{c}{AUC@30$^\circ$ $\uparrow$} \\
\cmidrule(lr){2-6}
& KITTI & NuScenes & Waymo & OpenScene & DDAD \\
\midrule
OmniVGGT    & 88.9 & 38.5 & 44.7 & 33.9 & 34.5 \\
CUT3R       & 51.8 & 43.5 & 50.1 & 34.7 & 48.6 \\
VGGT        & \underline{96.9} & \underline{87.8} & \underline{87.7} & 66.3 & 92.8 \\
MapAnything & 90.6 & 85.0 & 82.8 & 65.6 & 87.0 \\
StreamVGGT  & 95.8 & 86.2 & 85.6 & \underline{74.1} & 91.9 \\
DVGT        & 87.6 & 86.5 & 86.4 & \textbf{74.7} & \underline{95.1} \\
\midrule
\textbf{TRIG (Ours)} & \textbf{99.1} & \textbf{96.6} & \textbf{97.7} & 69.5 & \textbf{97.4} \\
\bottomrule
\end{tabular}
\end{table}

\paragraph{Rig supervision.}
\lizhou{Rig supervision constrains the relative geometry between cameras at the same timestamp.
For a camera pair $(c_i,c_j)$ at time $t$, the predicted and ground-truth relative transformations are computed as
\begin{equation}
\begin{aligned}
\hat{T}^{(k),t}_{c_i\leftarrow c_j}
&=
\hat{T}^{(k),t,c_i}_{w\rightarrow c}
\left(
\hat{T}^{(k),t,c_j}_{w\rightarrow c}
\right)^{-1}, \\
T^{t}_{c_i\leftarrow c_j}
&=
T^{t,c_i}_{w\rightarrow c}
\left(
T^{t,c_j}_{w\rightarrow c}
\right)^{-1}.
\end{aligned}
\end{equation}
Given that cameras share an identical ego pose at each timestamp, this term primarily constrains the fixed rig layout and promotes cross-camera consistency.}

\paragraph{Trajectory supervision.}
\lizhou{Trajectory supervision models ego-motion along each camera stream.
For camera $c$ and timestamps $(t_i,t_j)$, we define
\begin{equation}
\begin{aligned}
\hat{T}^{(k),c}_{t_i\leftarrow t_j}
&= \hat{T}^{(k),t_i,c}_{w\rightarrow c}
\left( \hat{T}^{(k),t_j,c}_{w\rightarrow c} \right)^{-1}, \\
T^{c}_{t_i\leftarrow t_j}
&= T^{t_i,c}_{w\rightarrow c}
\left( T^{t_j,c}_{w\rightarrow c} \right)^{-1}.
\end{aligned}
\end{equation}
With fixed camera identity, this relative transformation mainly reflects ego-vehicle motion across time.
For both rig and trajectory pairs, we adopt the relative pose loss from $\pi^3$~\cite{wang2025pi}, which combines translation regression and rotation geodesic error on $SO(3)$.
Unlike $\pi^3$'s unified relative pose supervision, TRIG partitions relative pairs according to the trajectory-rig decomposition, yielding $L_{\mathrm{rig}}$ for cross-camera geometry and $L_{\mathrm{traj}}$ for ego-motion.}

\lizhou{Following VGGT, we also supervise the focal components in the predicted pose encoding and apply all pose losses to each refinement stage with stage-dependent weights.
The final pose objective combines the rig, trajectory, and focal losses.
This decoupled pose supervision explicitly separates cross-camera rig constraints from temporal ego-motion constraints, improving geometric consistency over unified pose supervision.}

\begin{table}[t]
\centering
\caption{\textbf{Ablation studies on TRIG.}
All metrics are averaged over DDAD, KITTI, NuScenes, and Waymo.}
\label{tab:ablation}
\small
\setlength{\tabcolsep}{4pt}
\renewcommand{\arraystretch}{0.85}
\begin{tabular*}{\columnwidth}{@{\extracolsep{\fill}}lcccc@{}}
\toprule
Variant
& Abs Rel $\downarrow$
& $\delta_{1.25}$ $\uparrow$
& Acc. $\downarrow$
& Comp. $\downarrow$ \\
\midrule

\textit{Full model (Ours)}
& \textbf{0.069}
& \textbf{0.943}
& \textbf{0.411}
& \textbf{0.577} \\

\midrule
\multicolumn{5}{@{}l}{\emph{Training loss}} \\
VGGT pose loss
& 0.076
& 0.923
& 0.447
& 0.722 \\

\midrule
\multicolumn{5}{@{}l}{\emph{Geometry attention}} \\
w/o Rig block
& 0.163
& 0.871
& 0.881
& 0.908 \\
Frame-only
& 0.167
& 0.863
& 0.907
& 0.924 \\
Global-only
& 0.155
& 0.874
& 0.834
& 0.880 \\
3-layer rig block
& 0.113
& 0.901
& 0.606
& 0.776 \\

\midrule
\multicolumn{5}{@{}l}{\emph{Pose prior}} \\
OmniVGGT pose prior
& 0.090
& 0.924
& 0.533
& 0.691 \\

\bottomrule
\end{tabular*}
\end{table}

\begin{table}[h]
\centering
\caption{\textbf{Runtime comparison.}
Measured on NuScenes using a single NVIDIA L20 GPU.}
\label{tab:runtime}
\small
\setlength{\tabcolsep}{5pt}
\renewcommand{\arraystretch}{0.85}
\begin{tabular*}{\columnwidth}{@{\extracolsep{\fill}}lcccc@{}}
\toprule
Cams $\times$ Steps
& Frames
& Ours
& VGGT-style attn.
& Speedup $\uparrow$ \\
\midrule
$6 \times 21$
& 126
& \textbf{10.65s}
& 15.18s
& \textbf{1.43$\times$} \\
$6 \times 40$
& 240
& \textbf{24.87s}
& 42.86s
& \textbf{1.72$\times$} \\
\bottomrule
\end{tabular*}
\end{table}

\begin{table}[h]
\centering
\caption{\textbf{Pose ablation under multiple angular thresholds.}
All AUC scores are averaged over DDAD, KITTI, NuScenes, and Waymo.}
\label{tab:ablation_pose}
\small
\renewcommand{\arraystretch}{0.85}
\begin{tabular*}{\columnwidth}{@{\extracolsep{\fill}}lcccc@{}}
\toprule
\multirow{2}{*}{Variant}
& \multicolumn{4}{c}{Pose AUC $\uparrow$} \\
\cmidrule(lr){2-5}
& @5$^\circ$ & @10$^\circ$ & @15$^\circ$ & @30$^\circ$ \\
\midrule
\textit{Full model (Ours)}
& \textbf{86.92} & \textbf{92.56} & \textbf{94.77} & \textbf{97.69} \\
\midrule
\multicolumn{5}{@{}l}{\emph{Training loss}} \\
VGGT pose loss
& 76.33
& 85.58
& 89.53
& 94.19 \\
\midrule
\multicolumn{5}{@{}l}{\emph{Geometry attention}} \\
w/o Rig block
& 85.59 & 91.40 & 93.81 & 96.63 \\
Frame-only
& 86.02 & 91.78 & 94.13 & 96.57 \\
Global-only
& 86.07 & 91.73 & 94.07 & 96.71 \\
3-layer rig block
& 85.54 & 91.47 & 93.90 & 96.70 \\
\midrule
\multicolumn{5}{@{}l}{\emph{Pose prior}} \\
OmniVGGT pose prior
& 84.65 & 88.90 & 92.50 & 96.40 \\
\bottomrule
\end{tabular*}
\end{table}

\begin{table*}[t]
\centering
\caption{\textbf{Dataset-Specific Attention Ablation.} Performance breakdown of the Sparse Temporal--Spatial Attention (STSA) module. We report metric depth (Rel = Abs Rel, $\delta = \delta_{1.25}$), 3D reconstruction (Acc, Comp), and ego-pose (AUC = AUC@30$^\circ$) across four driving datasets. Best results are in \textbf{bold}.}
\label{tab:attn_ablation}
\renewcommand{\arraystretch}{1.15}
\setlength{\tabcolsep}{4pt}
\resizebox{\textwidth}{!}{
\begin{tabular}{@{}l ccccc ccccc ccccc ccccc@{}}
\toprule
\multirow{2}{*}{Variant}
& \multicolumn{5}{c}{DDAD}
& \multicolumn{5}{c}{KITTI}
& \multicolumn{5}{c}{NuScenes}
& \multicolumn{5}{c}{Waymo} \\
\cmidrule(lr){2-6} \cmidrule(lr){7-11} \cmidrule(lr){12-16} \cmidrule(lr){17-21}
& Rel$\downarrow$ & $\delta\!\uparrow$ & Acc$\downarrow$ & Comp$\downarrow$ & AUC$\uparrow$
& Rel$\downarrow$ & $\delta\!\uparrow$ & Acc$\downarrow$ & Comp$\downarrow$ & AUC$\uparrow$
& Rel$\downarrow$ & $\delta\!\uparrow$ & Acc$\downarrow$ & Comp$\downarrow$ & AUC$\uparrow$
& Rel$\downarrow$ & $\delta\!\uparrow$ & Acc$\downarrow$ & Comp$\downarrow$ & AUC$\uparrow$ \\
\midrule
\emph{Full model (Ours)}  & \textbf{0.099} & \textbf{0.915} & \textbf{0.559} & \textbf{0.767} & \textbf{98.85} & \textbf{0.041} & \textbf{0.977} & \textbf{0.291} & \textbf{0.306} & \textbf{99.44} & \textbf{0.051} & \textbf{0.974} & \textbf{0.317} & \textbf{0.356} & \textbf{96.29} & \textbf{0.085} & \textbf{0.906} & \textbf{0.478} & \textbf{0.879} & \textbf{96.19} \\
w/o Rig block             & 0.157 & 0.888 & 0.839 & 0.894 & 97.41 & 0.115 & 0.953 & 0.630 & 0.532 & 99.37 & 0.248 & 0.757 & 1.256 & 1.133 & 95.21 & 0.133 & 0.887 & 0.800 & 1.073 & 94.53 \\
Frame-only                & 0.157 & 0.886 & 0.844 & 0.898 & 96.19 & 0.113 & 0.953 & 0.627 & 0.529 & 99.34 & 0.260 & 0.733 & 1.329 & 1.179 & 95.47 & 0.137 & 0.882 & 0.826 & 1.089 & 95.27 \\
Global-only               & 0.146 & 0.900 & 0.773 & 0.875 & 97.36 & 0.106 & 0.955 & 0.589 & 0.510 & 99.14 & 0.248 & 0.743 & 1.268 & 1.128 & 95.31 & 0.120 & 0.899 & 0.706 & 1.006 & 95.02 \\
3-layer rig block         & 0.122 & 0.901 & 0.636 & 0.792 & 97.37 & 0.089 & 0.951 & 0.438 & 0.475 & 99.08 & 0.145 & 0.857 & 0.717 & 0.901 & 95.28 & 0.097 & 0.895 & 0.634 & 0.934 & 95.05 \\
\bottomrule
\end{tabular}
}
\end{table*}




\section{Experiments}
\lizhou{We conduct extensive experiments to evaluate the performance of TRIG. To ensure fair comparison, we follow the experimental setup of DVGT, including benchmark selection and evaluation protocols. Unless otherwise specified, baseline results are taken from the DVGT paper, and TRIG is evaluated under the same setting. Detailed dataset statistics, implementation hyperparameters, and formal metric definitions are provided in the Appendix.}

\subsection{3D Reconstruction and Metric Depth}
\noindent\textbf{3D Reconstruction.}
\lizhou{As summarized in Table~\ref{tab:recon}, TRIG establishes a new state-of-the-art in world-coordinate 3D Accuracy and Completeness across all five benchmarks. It consistently outperforms both general visual geometry models (e.g., CUT3R~\cite{wang2025continuous}, VGGT~\cite{wang2025vggt}) and driving-specific architectures (e.g., DVGT~\cite{zuo2026dvgt}). This quantitative dominance is visually corroborated in Fig.~\ref{fig:qualitative_results}, where TRIG yields significantly sharper geometries and cleaner boundaries than DVGT.}

\lizhou{Crucially, unlike image-only foundation models that require post-hoc Umeyama~\cite{umeyama1991least} alignment to recover scale, TRIG leverages metric pose priors to define a deterministic scale reference, allowing its normalized predictions to be directly restored to metric scale without additional alignment. While recent prior-guided models (OmniVGGT~\cite{peng2026omnivggt}, MapAnything~\cite{keetha2026mapanything}) attempt similar metric-aware predictions, they degrade severely in complex multi-camera setups (e.g., MapAnything's 4.303 Completeness error on OpenScene). This exposes the vulnerability of entangled priors in wide-baseline surround-view configurations with impoverished appearance cues. By routing temporal ego-motion and static camera-rig topology through isolated pathways, TRIG constructs a stable metric reference frame, ensuring geometric stability where purely visual evidence falls short.}

\noindent\textbf{Metric Depth.}
\lizhou{Table~\ref{tab:depth} demonstrates that the structural advantages of our decoupled representation seamlessly transfer to dense depth estimation. TRIG dominates the primary metric error (Abs Rel) and inlier ratio ($\delta_{1.25}$), achieving substantial margins over DVGT (the strongest prior-free baseline) on highly dynamic datasets like KITTI and NuScenes.}

\lizhou{On Waymo, while TRIG yields a marginally lower strict $\delta_{1.25}$ than DVGT, it maintains a superior global Abs Rel. Because the hard-threshold $\delta_{1.25}$ is disproportionately sensitive to isolated boundary artifacts, our consistently lower overall Abs Rel and 3D Completeness (Table~\ref{tab:recon}) firmly prove that TRIG achieves a robust global metric consensus rather than overfitting to specific local thresholds. Ultimately, this joint improvement in depth and reconstruction confirms that decoupling dynamic and static poses is a potent geometric catalyst.}

\noindent\textbf{Ego-Pose Estimation.}
\lizhou{Following VGGT~\cite{wang2025vggt}, we evaluate relative camera pose accuracy using AUC@$30^\circ$. Table~\ref{tab:pose} shows TRIG secures the highest pose accuracy across KITTI, NuScenes, Waymo, and DDAD. Compared to DVGT, TRIG yields profound improvements (e.g., elevating KITTI from 87.6 to 99.1). Furthermore, TRIG circumvents the instability exhibited by recent prior-guided models. Whereas entangled prior injection causes OmniVGGT to degrade severely under wide-baseline configurations—dropping to 38.5 AUC on NuScenes—TRIG maintains a substantially higher AUC of 96.6.}

\lizhou{These gains are critical for autonomous driving, where drifting poses render downstream reasoning unusable. Rooted in our decoupled supervision strategy, TRIG isolates inter-frame temporal constraints (ego-trajectory) from intra-frame spatial constraints (camera-rig). This explicitly optimizes ego-motion without corrupting the static surround-view layout. Consequently, TRIG delivers highly accurate, drift-resistant metric poses, ensuring the reconstructed scene is directly actionable in a driving-oriented 3D coordinate system.}

\subsection{Ablation Studies}
\label{sec:ablation}
\lizhou{We conduct extensive ablations across the DDAD, KITTI, NuScenes, and Waymo benchmarks to validate TRIG's core architectural propositions. As summarized in Tables~\ref{tab:ablation}, ~\ref{tab:ablation_pose} and ~\ref{tab:attn_ablation}, we jointly assess metric depth, 3D reconstruction, and ego-pose estimation to capture holistic geometric impacts.}

\noindent\textbf{Sparse Temporal--Spatial Attention.}
\lizhou{We evaluate Sparse Temporal--Spatial Attention (STSA) with four variants: \textit{w/o Rig block}, \textit{Frame-only}, \textit{Global-only}, and \textit{3-layer rig block}, where Rig blocks are inserted at layers 0, 11, and 23. As shown in Table~\ref{tab:ablation}, removing Rig blocks increases the reconstruction Accuracy error from 0.411 to 0.881. \textit{Frame-only} performs even worse, suggesting that local refinement alone is insufficient for preserving global geometry. Although \textit{Global-only} and \textit{3-layer rig block} improve upon these weaker baselines, both remain inferior to full STSA, highlighting the importance of jointly modeling temporal and cross-view interactions. Table~\ref{tab:attn_ablation} further demonstrates consistent improvements across DDAD, KITTI, NuScenes, and Waymo. In addition, Table~\ref{tab:runtime} shows that STSA is more efficient than VGGT-style attention, achieving $1.43\times$ and $1.72\times$ speedups on $6{\times}21$ and $6{\times}40$ NuScenes inputs.}

\noindent\textbf{Decoupled Prior Encoding.}
\lizhou{To demonstrate the superiority of our decoupled strategy, we replace our distinct trajectory and rig embeddings with the entangled OmniVGGT pose prior. Table~\ref{tab:ablation} shows marked deterioration across all metrics, with point completeness worsening from 0.577 to 0.691. These results provide empirical support for our central hypothesis: ego-motion and camera-rig calibration represent intrinsically heterogeneous sources of information that warrant separate modeling. By routing them through decoupled pathways, TRIG smoothly integrates multi-modal geometric constraints without corrupting the foundation model's original feature space.}

\noindent\textbf{Decoupled Pose Supervision.}
\lizhou{To assess our decoupled pose supervision, we replace it with the standard VGGT's pose loss. As shown in Tables~\ref{tab:ablation} and~\ref{tab:ablation_pose}, this replacement consistently degrades both pose accuracy and geometric quality. Our full model substantially improves Pose AUC under all angular thresholds, especially from 76.33 to 86.92 at $5^\circ$, indicating stronger resistance to pose drift. Meanwhile, it also improves dense depth and reconstruction, reducing Abs Rel from 0.076 to 0.069 and Completeness error from 0.722 to 0.577. These results indicate that separately supervising temporal ego-motion and spatial rig constraints is more effective than using a VGGT pose loss.}



\section{Conclusion}

\lizhou{In this paper, we introduced TRIG, a novel metric geometry perception framework for multi-camera autonomous driving. By explicitly decoupling dynamic ego-trajectories from static camera-rig topologies, TRIG overcomes the geometric entanglement and scale ambiguity inherent in existing visual foundation models. To achieve this, we proposed a decoupled prior encoding module, a highly efficient Sparse Temporal--Spatial Attention (STSA) mechanism, and a decoupled pose supervision strategy. Extensive experiments across five large-scale driving benchmarks demonstrate that TRIG establishes new state-of-the-art standards in metric depth prediction, 3D scene reconstruction, and ego-pose estimation, providing a robust, scale-deterministic solution for end-to-end 3D scene understanding.}

\appendix \label{Appendix}
\setcounter{table}{0}
\setcounter{figure}{0}
\setcounter{equation}{0}
\renewcommand{\thetable}{\Alph{section}.\arabic{table}}
\renewcommand{\thefigure}{\Alph{section}.\arabic{figure}}
\renewcommand{\theequation}{\Alph{section}.\arabic{equation}}

\section{Dataset Details}
We train our model on 5 public autonomous driving datasets. All datasets provide calibrated camera intrinsics and extrinsics, along with ground-truth depth and pose annotations. Table~\ref{tab:dataset_stats} summarizes the key statistics of each dataset.

\begin{figure*}[!t]
    \centering
    \includegraphics[width=0.48\textwidth]{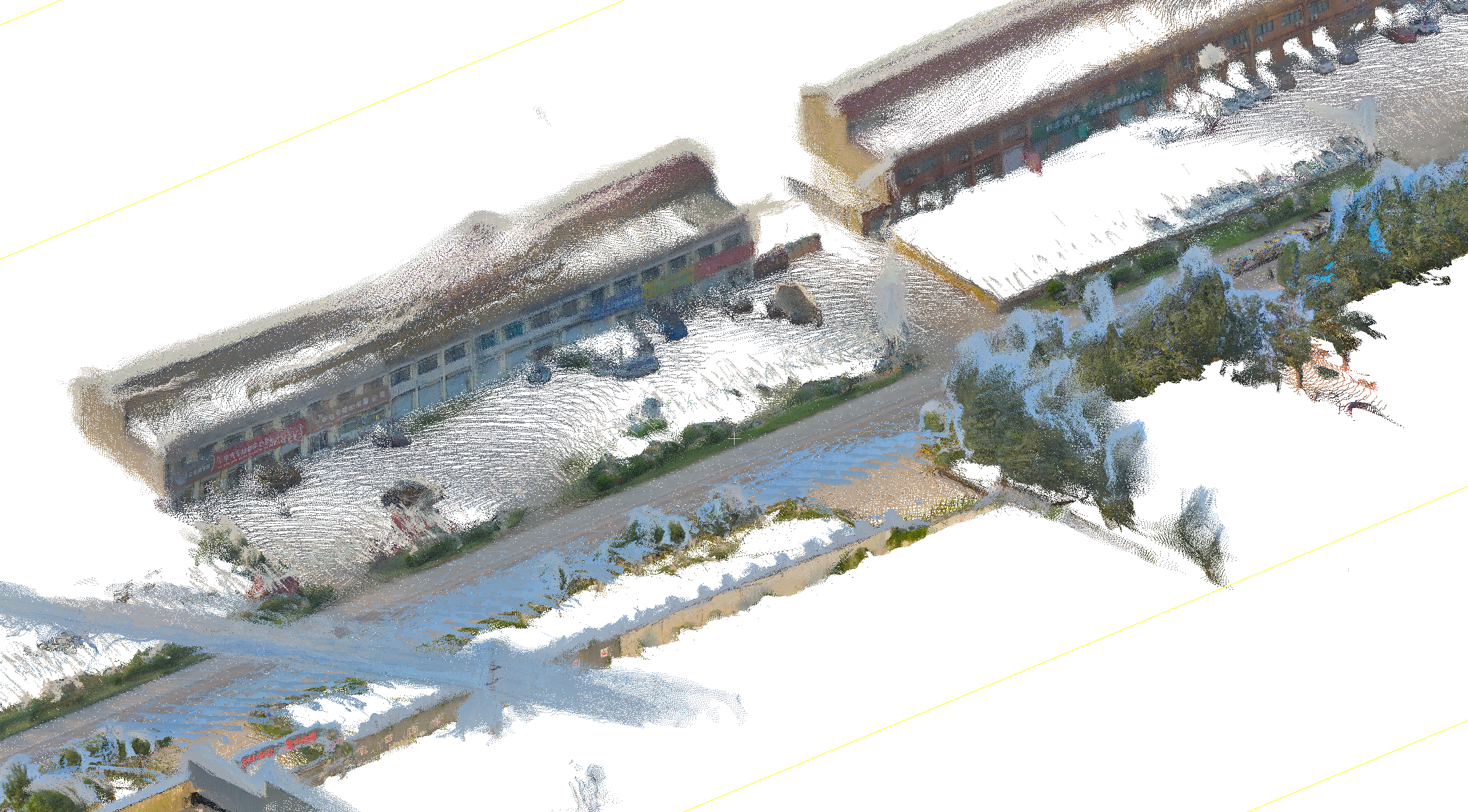}
    \hfill
    \includegraphics[width=0.48\textwidth]{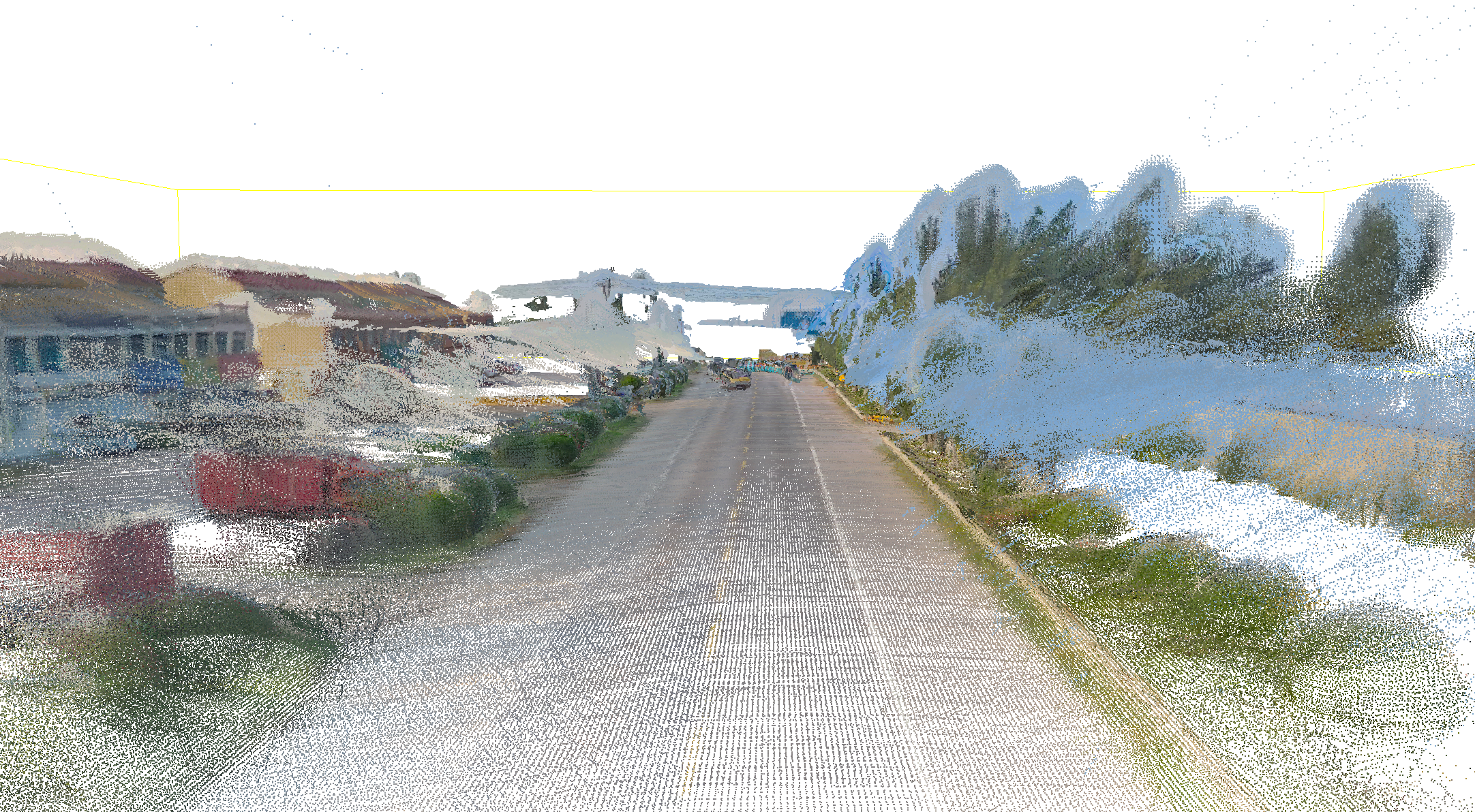}
    \vspace{0.6em}
    \includegraphics[width=0.48\textwidth]{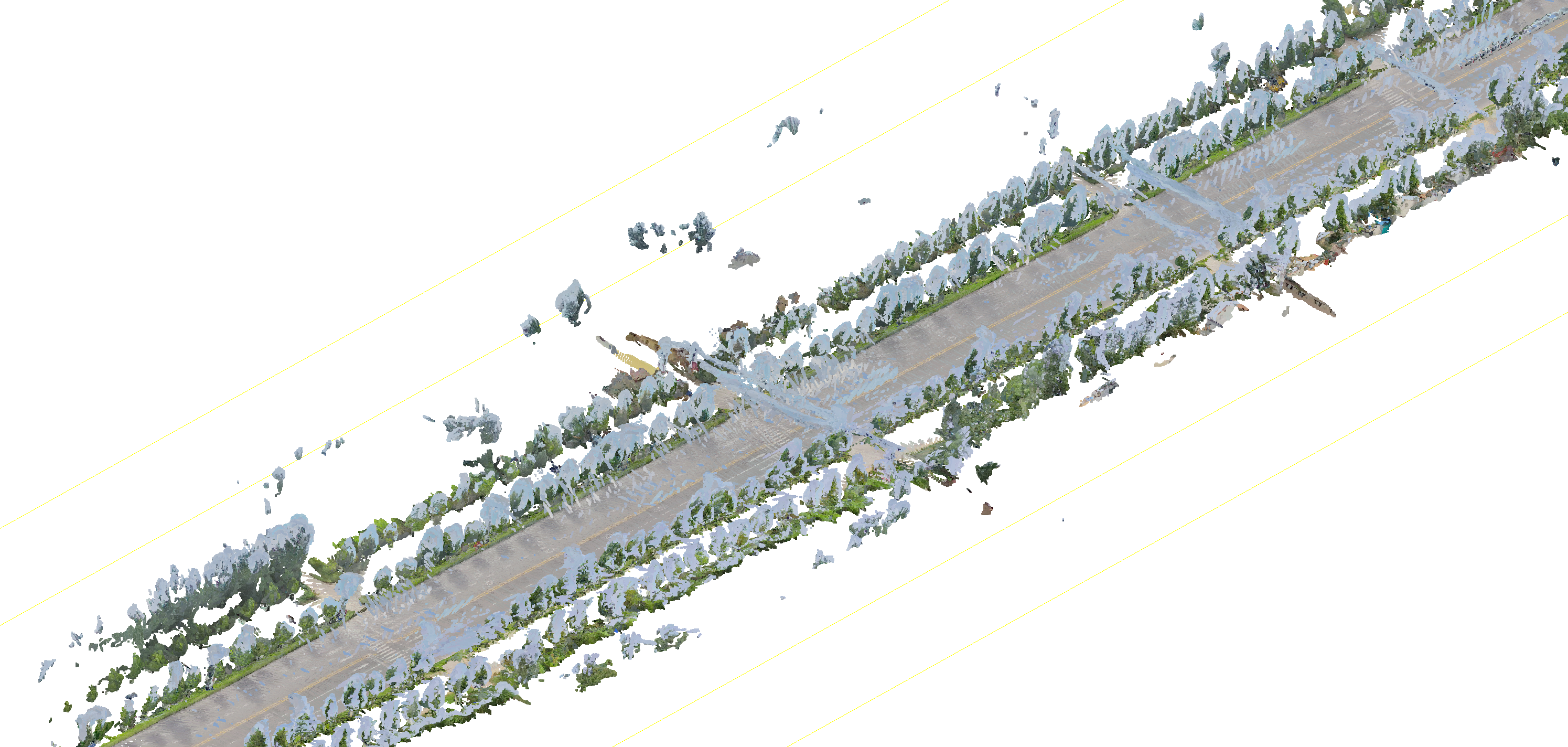}
    \hfill
    \includegraphics[width=0.48\textwidth]{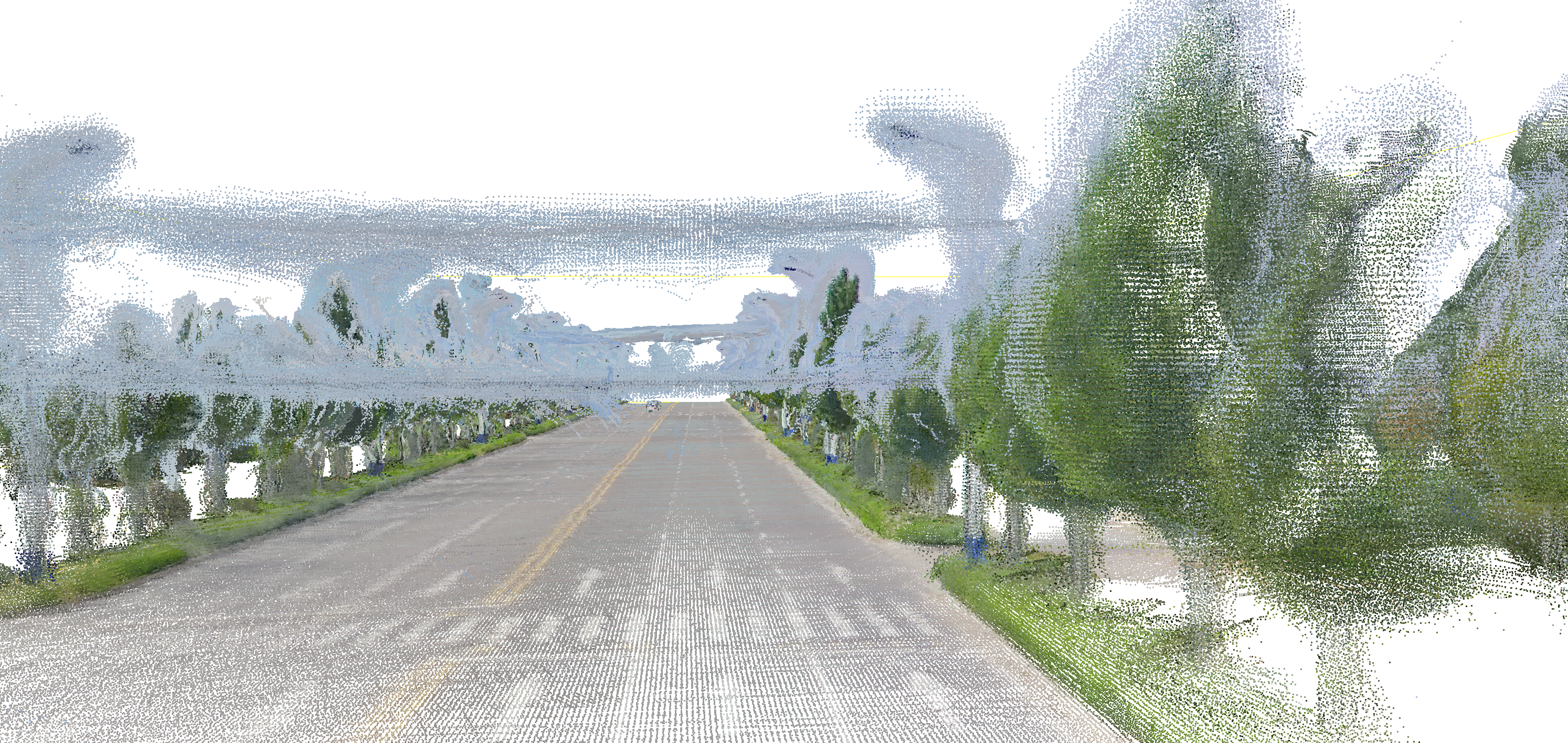}
    \vspace{0.6em}
    \includegraphics[width=0.48\textwidth]{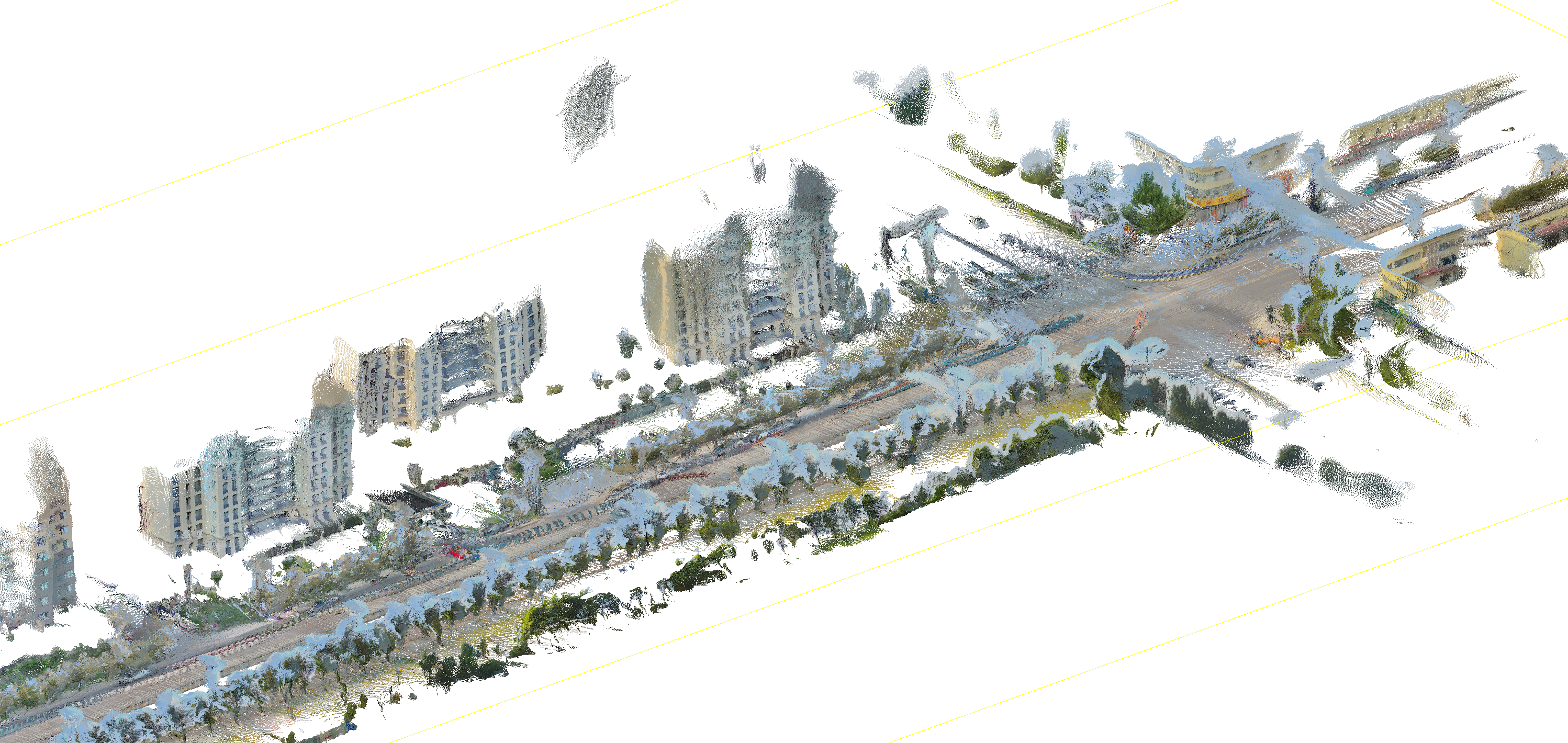}
    \hfill
    \includegraphics[width=0.48\textwidth]{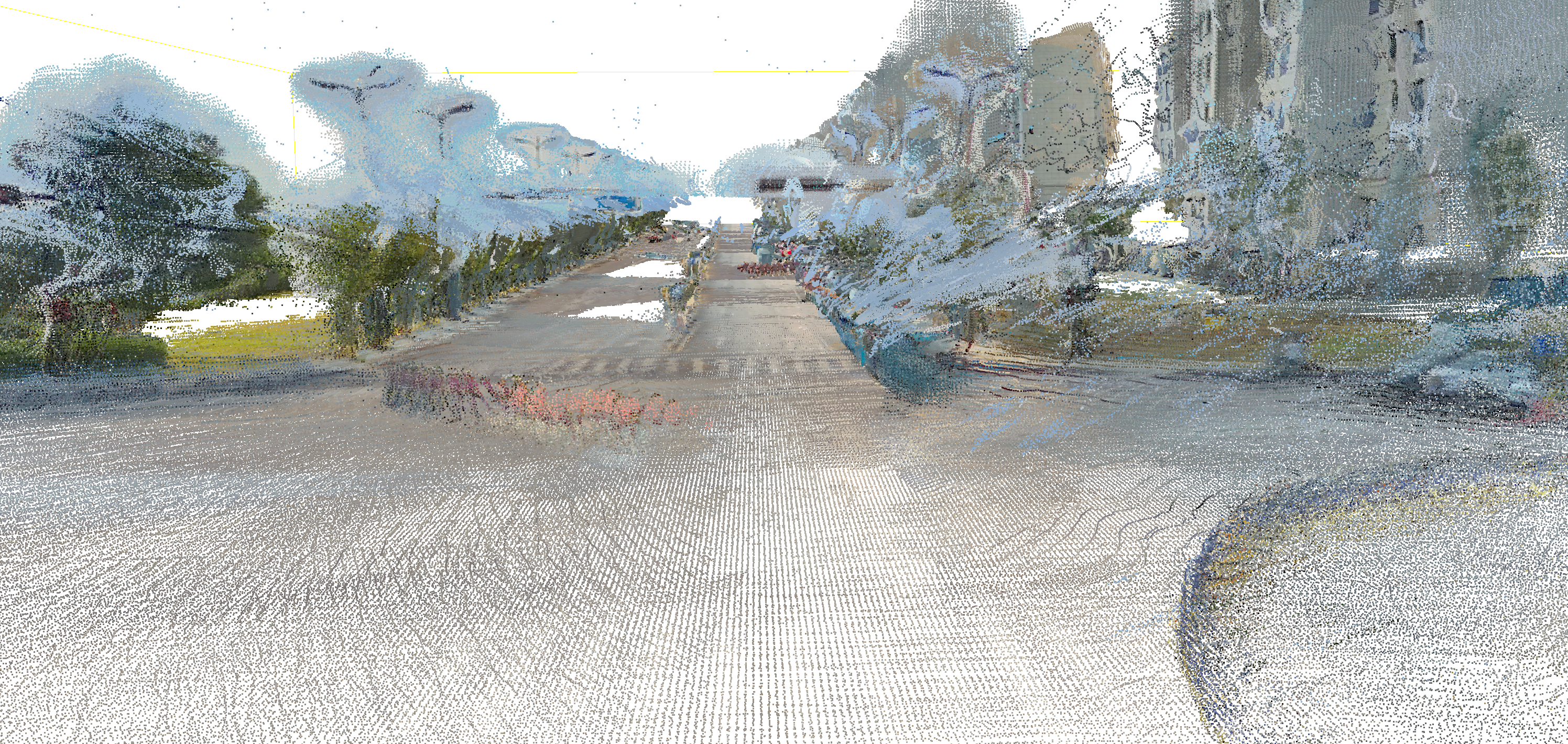}
    \caption{Reconstruction results on real-world autonomous driving sequences captured by a vehicle equipped with six cameras, with each sequence spanning 30--40 seconds. Predictions are rescaled by the pose-normalization scale factor without Sim(3) alignment, and the final point clouds are voxel-filtered for visualization.}
    \label{fig:real_drive_recon}
\end{figure*}

\section{Additional Implementation Details}

\subsection{Training Objective}

The full training objective of TRIG consists of three components: a depth loss, a point map loss, and the Trajectory--Rig Decoupled pose loss:
\begin{equation}
\mathcal{L}
=
\mathcal{L}_{\mathrm{depth}}
+
\mathcal{L}_{\mathrm{pmap}}
+
\mathcal{L}_{\mathrm{pose}},
\end{equation}
where the pose loss is further decomposed into a trajectory loss and a rig loss:
\begin{equation}
\mathcal{L}_{\mathrm{pose}}
=
\mathcal{L}_{\mathrm{traj}}
+
\mathcal{L}_{\mathrm{rig}}.
\end{equation}
The trajectory loss supervises the inter-frame relative poses of the front camera, while the rig loss supervises the inter-camera relative poses at the reference timestamp $t{=}0$. This decoupled formulation encourages the model to separately learn temporal ego-motion and static multi-camera rig geometry, reducing the ambiguity between vehicle motion and camera extrinsics.

\subsection{Model Architecture}

TRIG adopts a ViT-L backbone pretrained with DINOv2 as the image encoder. The image encoder is kept frozen during training to preserve the general-purpose visual representation learned from large-scale pretraining and to reduce the optimization cost. Given an input sequence with $T$ timestamps and $N_C$ cameras, the images are first encoded into visual tokens by the frozen ViT-L encoder.

To inject explicit pose-aware structure into the network, we separately encode the trajectory pose and the rig pose using two linear projection layers. Both projections map their corresponding pose representations into a 1024-dimensional latent space. The trajectory embedding has shape $(T, 1, 1024)$, while the rig embedding has shape $(1, N_C, 1024)$. We then combine them through broadcasting and summation:
\begin{equation}
\mathbf{E}_{\mathrm{pose}}
=
\mathbf{E}_{\mathrm{traj}}
+
\mathbf{E}_{\mathrm{rig}},
\end{equation}
resulting in a pose embedding of shape $(T, N_C, 1024)$. This embedding provides each camera-view token with both temporal trajectory information and camera-specific rig information.

On top of the image encoder, we employ the proposed Sparse Temporal--Spatial Attention (STSA) module for efficient feature interaction across time and camera views. STSA consists of two types of transformer blocks:

\begin{itemize}
    \item \textbf{Trajectory Block.}
    The trajectory block is the main building block of STSA and is repeated for 24 layers. Each trajectory block first applies frame-level self-attention to aggregate information within each timestamp, and then performs per-camera temporal attention to model ego-motion over time. This design captures temporal dynamics efficiently while avoiding the quadratic cost of full spatio-temporal attention over all frames and cameras.

    \item \textbf{Rig Block.}
    To model cross-camera geometry, we insert rig blocks at selected layers, specifically at indices $\{0,4,11,17,23\}$. Each rig block performs global cross-view attention across cameras, enabling the network to aggregate multi-camera spatial information and reason about the static camera rig configuration.
\end{itemize}

During STSA processing, the pose prior is injected into the frame-level attention of each layer, following the spirit of OmniVGGT. This allows the model to condition feature interaction on the decoupled trajectory and rig embeddings, thereby improving its ability to distinguish temporal motion from multi-camera geometric structure.

For dense prediction, we follow the design of VGGT and feed intermediate tokens from the 4th, 11th, 17th, and 23rd transformer blocks into a DPT decoder for upsampling. The DPT decoder progressively fuses multi-scale transformer features and produces dense outputs such as depth maps and point maps.

\begin{table}[t]
\centering
\caption{\textbf{Training dataset statistics.} All datasets are sampled at 2\,Hz and resized to $960{\times}540$ (Waymo: $576{\times}384$). Model input resolution is $518{\times}294$.}
\label{tab:dataset_stats}
\setlength{\tabcolsep}{3pt}
\renewcommand{\arraystretch}{0.9}
\footnotesize
\begin{tabular}{@{}l c c c c @{}}
\toprule
Dataset & \# Cameras & \# Frames & Original Res.  & Weight \\
\midrule
NuScenes    & 6 & $\sim$22K  & $1600{\times}900$  & 6  \\
Waymo       & 5 & $\sim$157K & $1920{\times}1280$ & 6  \\
KITTI       & 2 & $\sim$20K  & $1242{\times}375$  & 5  \\
DDAD        & 6 & $\sim$13K  & $1936{\times}1216$ & 6  \\
OpenScene   & 8 & $\sim$513K & $1920{\times}1080$ & 77 \\
\bottomrule
\end{tabular}
\end{table}

\subsection{Training Details}

TRIG is trained on multiple public datasets for 40K iterations. We follow the data preprocessing protocol of DVGT. All input images are resized to $518 \times 294$, and the patch size of the ViT encoder is $14 \times 14$.

We use AdamW as the optimizer with a base learning rate of $1 \times 10^{-4}$, weight decay of $5 \times 10^{-2}$, and betas of $(0.9, 0.95)$. Bias and normalization parameters are excluded from weight decay. The learning rate is scheduled by OneCycleLR with cosine annealing, where the warm-up phase takes the first $10\%$ of training iterations. The initial and final learning-rate division factors are both set to 100.

Training is performed with DeepSpeed ZeRO-1 optimization and bfloat16 mixed precision. We apply gradient norm clipping with a threshold of 1.0 for stable optimization. The model is trained on 64 NVIDIA H20 GPUs for approximately six days.

\section{Evaluation Metrics}

We evaluate our method's performance on three primary tasks: 3D point map reconstruction, metric depth estimation, and ego-pose estimation.

\textbf{3D Point Map Reconstruction.}
Following prior works, we measure overall geometric quality using \emph{Accuracy} and \emph{Completeness}.
Let $\mathcal{P} = \{p_i\}$ and $\mathcal{G} = \{g_j\}$ denote the sets of valid points from the predicted point map $\hat{P}$ and the ground truth $P$, respectively.
These metrics are calculated as:
\begin{align}
\mathrm{Accuracy}   &= \frac{1}{|\mathcal{P}|} \sum_{p \in \mathcal{P}} \min_{g \in \mathcal{G}} \|p - g\|_2, \label{eq:acc} \\
\mathrm{Completeness} &= \frac{1}{|\mathcal{G}|} \sum_{g \in \mathcal{G}} \min_{p \in \mathcal{P}} \|g - p\|_2, \label{eq:comp}
\end{align}
where $\|\cdot\|_2$ denotes the $L_2$ distance.
Accuracy measures how close each predicted point is to the nearest ground-truth point, reflecting reconstruction precision; Completeness measures how well the ground-truth geometry is covered by the prediction, indicating reconstruction recall.
Both metrics are evaluated in the world coordinate frame at metric scale, without any post-hoc Sim(3) alignment.

\textbf{Metric Depth Estimation.}
We evaluate per-camera depth using the \emph{Absolute Relative Error (Abs Rel)} and the \emph{inlier ratio $\delta_{1.25}$}.
Let $\Omega$ be the set of valid pixels with available ground truth, and let $u$ index a pixel within $\Omega$.
Given the predicted metric depth $D(u)$ and ground truth $D_{\mathrm{gt}}(u)$, the metrics are defined as:
\begin{align}
\mathrm{Abs\ Rel} &= \frac{1}{|\Omega|} \sum_{u \in \Omega} \frac{|D(u) - D_{\mathrm{gt}}(u)|}{D_{\mathrm{gt}}(u)}, \label{eq:absrel} \\
\mathrm{Acc}_\delta &= \frac{1}{|\Omega|} \sum_{u \in \Omega} \mathbb{I}\!\left(
\max\!\left( \frac{D(u)}{D_{\mathrm{gt}}(u)}, \frac{D_{\mathrm{gt}}(u)}{D(u)} \right) < \delta
\right), \label{eq:delta}
\end{align}
where $\mathbb{I}(\cdot)$ is the indicator function.
Abs Rel captures the average proportional error, while $\delta_{1.25}$ measures the fraction of pixels whose predicted depth falls within $25\%$ of the ground truth.

\textbf{Ego-Pose Estimation.}
Following VGGT, we report the \emph{Area Under the Curve (AUC)} at a $30^\circ$ threshold.
We first compute the Relative Rotation Accuracy (RRA) and Relative Translation Accuracy (RTA) for all frame pairs within a sequence.
RRA measures the geodesic distance between the predicted and ground-truth rotation matrices, while RTA evaluates the angular deviation between translation vectors.
The accuracy at threshold $\tau$ is the percentage of pairs satisfying $\max(\mathrm{RRA}, \mathrm{RTA}) < \tau$, and the final AUC@$30^\circ$ is obtained by integrating this accuracy:
\begin{equation}
\mathrm{AUC@30^\circ} = \frac{1}{30} \int_0^{30} \mathrm{Acc}(\tau) \, d\tau,
\label{eq:auc}
\end{equation}
where $\mathrm{Acc}(\tau)$ represents the fraction of camera pairs with both angular errors smaller than $\tau$.


\section*{Acknowledgments}

The authors thank Carizon colleagues for their valuable discussions, technical support, and constructive feedback during this work.

\bibliographystyle{IEEEtran}
\bibliography{references}

\end{document}